# When Shortest Isn't Safest: A Design Science Approach to Senior-Friendly Pedestrian Routing

Erdi Ünal[1][0009-0007-2809-030X], Daniel Eisenhardt[1][0009-0003-6114-6863], Christian Meske[1][0000-0001-5637-9433], Seyed Nima Afzali[2][0009-0002-4976-7918], Aysegül Dogangün[2][0000-0002-4287-3884]

[1] Ruhr University Bochum, Bochum, Germany
{erdi.uenal, daniel.eisenhardt, christian.meske}@rub.de
[2] Hochschule Ruhr West University of Applied Science, Bottrop, Germany
{ayseguel.doganguen, nima.afzali}@hs-ruhrwest.de

**Abstract.** Older adults' independent mobility enables out-of-home participation, well-being and health, yet pedestrian navigation systems still optimize primarily for distance or time, often overlooking barriers, safety thresholds, and supportive infrastructure that shape late-life walking decisions. We present a senior-friendly pedestrian routing artefact developed through echeloned Design Science Research, translating lived mobility constraints into prescriptive design knowledge. Based on 11 semi-structured interviews, we derive initial Design Requirements (DRs) and Design Principles (DPs) for barrier-aware, amenity-sensitive routing and execution-relevant explanations. We instantiate these in an OpenStreetMap pedestrian network enriched with amenities (benches, toilets, and shelters) and height data, and implemented an A*-based routing engine with configurable costs and explanation payloads. In a field-based walking study, 14 older adults compared artefact-generated routes with baselines and provided ratings and qualitative feedback; the senior-friendly route was preferred overall. Thematic analysis further showed that infrastructure maintenance, seasonal conditions, traffic exposure, and social context shape route acceptance. We synthesize these insights into refined DRs and DPs emphasizing context-aware hazard modeling, multi-route transparency, landmark-grounded explanations, social-context sensitivity, and stage-appropriate information. Our contributions provide actionable guidance for practitioners developing senior-friendly pedestrian navigation systems.

## 1 Introduction

Social participation is a fundamental human need, but it is particularly consequential in older age, where its absence negatively affects physical, cognitive, and mental health, as well as well-being [9, 14, 22] and increases mortality by 32–33% [20, 36]. At the same time, social isolation is becoming increasingly widespread, with 26% of community-dwelling older adults being socially isolated [29, 44]. As rising life expectancy, driven by advances in public health, nutrition, hygiene, and medicine, accelerates population ageing [32], supporting older adults' continued participation becomes increasingly important. Participation, however, requires being able to leave home and reach everyday destinations safely and confidently. Yet age-related declines in balance, gait stability, and physical capacity make walking increasingly effortful [11], contributing to restricted life-space mobility [13]. Technologies that reduce barriers to going out may therefore support well-being by helping older adults maintain participation and independence, goals older adults themselves strongly prioritize [24].

While pedestrian navigation systems support independent mobility, most mainstream solutions optimize primarily for path length or travel time, implicitly assuming homogeneous users and overlooking constraints such as stairs, uneven surfaces, fatigue, or rest opportunities [28]. For older adults, "shortest" can conflict with "safest" or "most manageable": their route choice depends on barrier-free infrastructure, perceived safety, comfort, and supportive amenities. This makes information about route difficulty essential for planning and with multi-criteria routing becoming more complex, communicating rationales behind recommendations ("longer but safer", "avoids stairs", "more benches") becomes essential for trust and informed decision-making [8, 35, 37].

Research addresses these limitations via multi-criteria and accessibility-aware pedestrian routing, integrating factors like slope, surface quality, obstacles, or amenities [28]. But much of this work is technically oriented, and senior-specific contributions are limited to prototypes without transparent explanation logic or field-based evaluation with older adults. Recently, three factor dimensions for walkable route design were synthesized (street-infrastructure and structural attributes; environmental quality and aesthetics; facility access and contextual features), reinforcing that route choice extends beyond time and distance [12]. Still, empirically grounded guidance for senior-friendly navigation, especially on how to communicate trade-offs and execution-relevant information, is scarce. This reflects broader shortcomings in IS research on technology for older adults: the neglect of biophysical and psychosocial factors, like physical limitations, fear, and confidence, that are among the strongest determinants of older adults' technology use [26]. Thus, prescriptive design knowledge is needed to guide practitioners in developing senior-oriented routing algorithms, leading to our research question.

RQ: *What design requirements and design principles inform barrier-sensitive pedestrian routing and execution-relevant route communication for older adults?*

To answer this, we follow an echeloned Design Science Research (eDSR) approach [31] to develop and evaluate a senior-friendly pedestrian routing artefact. We elicit Design Requirements (DRs) and Design Principles (DPs) via a participatory analysis with older adults and instantiate them in a routing engine and structured explanation outputs, using open pedestrian network data. We evaluate in multiple steps, culminating in a field-based comparison against a Google Maps baseline, enabling us to derive prescriptive design knowledge grounded in older adults' lived mobility constraints. Thus, we respond to calls for preventive digital artefacts that support aging populations outside formal healthcare settings, an underexplored IS research opportunity [3].

The paper proceeds as follows: Section 2 reviews literature on older adults' mobility constraints, route choice determinants, and senior-friendly pedestrian navigation; Section 3 describes the method; Section 4 presents the resulting design knowledge and artefact instantiation across two iterative cycles; Sections 5 and 6 discuss and conclude.

## 2 Background

### 2.1 The Impact of Mobility on Older Adults

Independent mobility describes the ability to leave one's home and reach everyday destinations safely and confidently [38], a central prerequisite for autonomy in life. With increasing age, declines in gait stability, balance, and physical capacity can make walking and wayfinding effortful and uncertain [11]. As a result, many older adults progressively restrict their outdoor activity range [13]. This reduction is consequential because mobility is tightly linked to maintaining regular physical activity, which is a key protective factor for healthy aging and functional independence [43]. Over time, these constraints can initiate a self-reinforcing cycle of reduced walking, deconditioning, and further mobility loss [41]. Beyond physical functioning, reduced independent mobility and social isolation can trigger broader psychosocial consequences. When older adults avoid going out due to fatigue, insecurity, or fear of falling, everyday encounters and opportunities for participation decrease. Regular social participation is essential for maintaining well-being, cognitive function, and overall health in later life [9, 14, 22]. Conversely, restricted mobility can contribute to social isolation, which represents a growing public health concern [44]. Social isolation is linked to increased mortality, stressing its significance for healthy aging [20, 36].

Importantly, these effects can reinforce each other in a self-amplifying cycle: mobility restrictions reduce physical activity and social engagement, which can accelerate deconditioning, lower confidence, and increase withdrawal, further limiting mobility over time. Supporting older adults in maintaining safe outdoor mobility may therefore contribute indirectly to multiple health domains by preserving participation and preventing avoidable decline. This motivates mobility-support technologies that reduce barriers to going out and lower uncertainty during trips. To design effective support technologies, we must understand which environmental factors shape older adults' walking decisions. The next section translates this motivation into design-relevant evidence by reviewing the key determinants of route choice and their prioritizations.

### 2.2 Environmental Determinants of Older Adults' Walking and Route Choice

Mobility in later life depends not only on personal factors but also on contextual conditions and person-environment interactions [42]. An established hierarchy of walking needs [1] distinguishes five levels of environmental relevance: feasibility, accessibility, safety, comfort, and pleasurability. Lower-order needs must be satisfied before higher-order preferences shape walking decisions. This provides practical design guidance: accessibility and safety serve as constraints, whereas comfort and aesthetics become optimization targets once basic requirements are fulfilled.

Empirical evidence on older adults' walking and route choice comes from diverse qualitative and quantitative approaches [12] which we synthesize in Table 1. Although much of this work was conducted in high-income contexts, and the salience of comfort-related features and specific trade-offs may therefore vary across climates and urban forms, the literature converges on a recurring set of route-choice factors. These include infrastructure and structural barriers, safety-related features, comfort and amenities, aesthetics and environmental quality, and navigation- and orientation-related aspects [12, 23]. In addition, neighborhood walkability moderates the relationship between physical functioning and transport walking, while perceived outdoor barriers can precede later walking difficulties [25, 34].

Taken together, the evidence supports representing older adults' pedestrian route choice as a process of sequential constraint satisfaction. Accessibility-related barriers act as hard constraints, safety features operate as threshold requirements, and comfort, amenities, and environmental quality enable personalized optimization once lower-order needs are met. Yet mainstream navigation applications rarely account for these layered constraints and preferences.

**Table 1.** Influential factors for older adults' walking and route choice.

| Factor Class | Key Determinants | Role in Route Choice | Refs |
|---|---|---|---|
| Infrastructure & barriers | Effective "walkable" sidewalk width, micro-scale irregularities (uneven pavement, curb height, steps/stairs, no handrails), clutter, discontinuities, shared paths, and gradients | Hard constraints; translate directly into fall risk and fatigue; intensified for individuals with mobility limitations or fear of falling; context-dependent (e.g., trip purpose, hilly areas) | [6, 15, 19, 33] |
| Safety (traffic & perceived) | Busy roads, fast traffic, complex crossings, poor lighting, sightlines, signs of disorder, "openness", visual surveillance | Threshold condition; avoided in principle but traded off against destination access | [4, 7, 18, 33] |
| Comfort & amenities | Benches, toilets, shelters, rest opportunities | Expands feasible walking range; importance moderated by health status | [6, 17, 18, 33] |
| Aesthetics & environmental quality | Greenery, cleanliness, maintenance, noise/air quality | Affect enjoyment once lower-order needs are met; greenery can backfire when reducing visibility or feeling isolating | [4, 6, 19] |
| Navigation & orientation | Signage, layout legibility, landmarks, sensory complexity | Constrain route choice in unfamiliar environments; landmarks support orientation | [23] |

### 2.3 Senior-Friendly Pedestrian Navigation

This gap reflects a broader pattern in IS research: technically functional systems deployed for older adults frequently fail when they conflict with users' emotional, physical, and social practices [39]. Addressing the mobility barriers identified above therefore requires both proactive route selection (route-level support) and real-time guidance during walks (in-route support). In response, pedestrian routing research has moved toward multi-criteria and accessibility-aware approaches that incorporate factors such as slope, surface quality, crossings, obstacles, and amenities into route calculation. Penalty-based methods, for example, increase the cost of undesirable segments such as stairs, steep ramps, or narrow footways to steer algorithms toward more accessible routes while remaining feasible for real-time use [28]. Such systems must often negotiate trade-offs, for example between shorter and safer or more manageable routes. This creates a need for increased transparent recommendations that help users understand why a route was chosen and what conditions to expect. Transparency is critical for trust and informed decision-making as multi-criteria routing becomes more complex [37]. This is especially relevant for older adults, who have been shown to value transparency about decision-making processes when interacting with AI-driven systems [8]. For older adults, interpretable route characteristics and clear communication of hazards, rest opportunities, and trade-offs may provide reassurance and support collaborative engagement with routing decisions [35].

Although older adults' route preferences have been studied, this evidence has only rarely been translated into senior-oriented navigation design. Existing work has explored personalized pedestrian routing [21, 27], comfortable senior routes based on weighted criteria such as incline and amenities [2], and accessibility-oriented routing models that account for features such as footway width, slope, curb design, tactile paving, and obstacles [16]. However, these systems are largely presented as technical prototypes, with limited evidence on how older adults understand, assess, and act on route recommendations. Thus, empirically grounded prescriptive knowledge for designing senior-friendly navigation systems remains scarce, motivating our research project.

## 3 Methodology

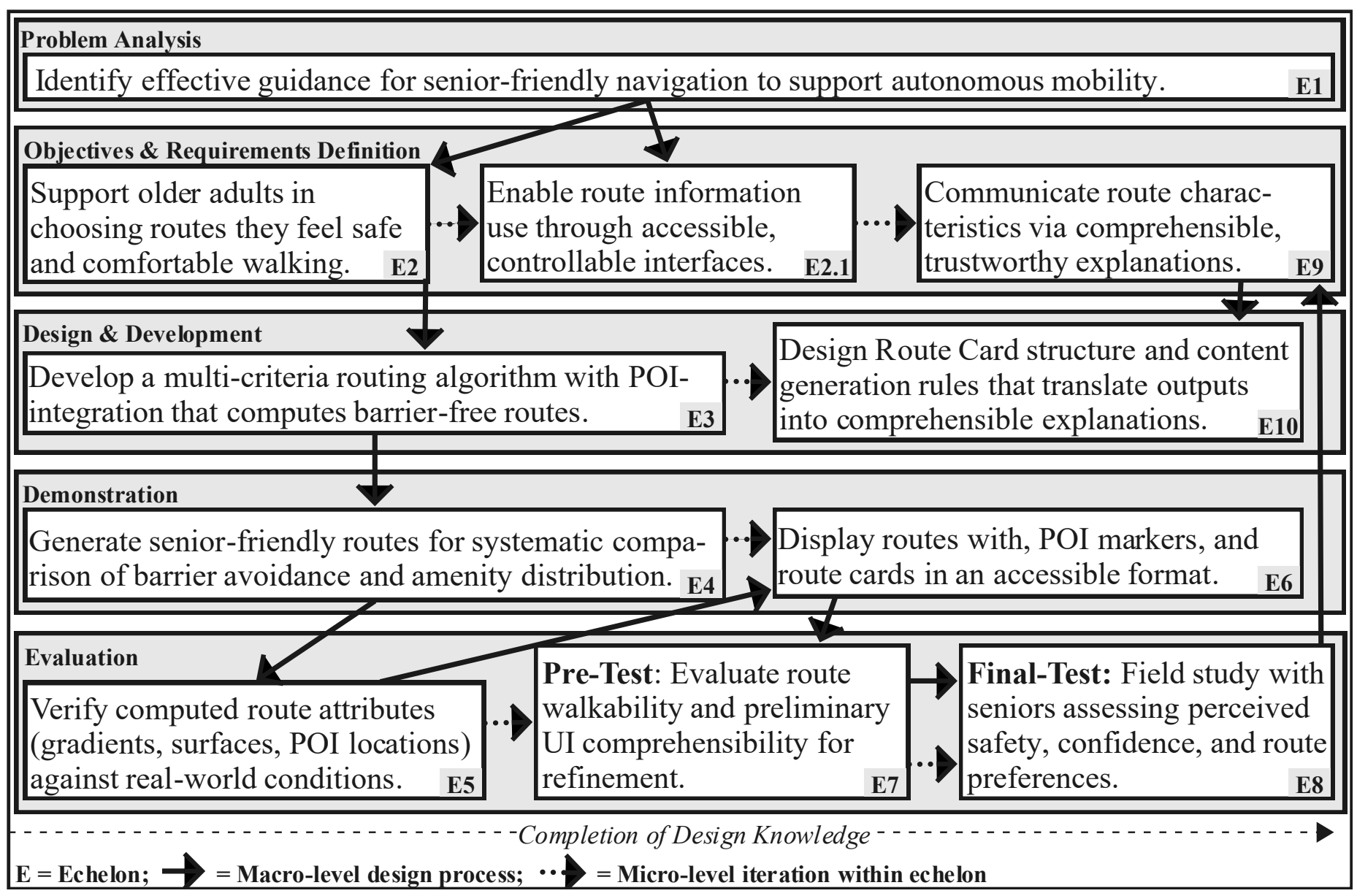

**Fig. 1.** Overview of the adapted eDSR process.

We adopted the eDSR methodology, allowing us to decompose complex design science efforts into interdependent subsystems ("echelons") that can be designed and evaluated iteratively while preserving coherence of the overall artefact [31]. Given the diversity and interdependence of older adults' requirements, this structure enabled us to analyze and address specific sub-goals in depth, reducing the risk of overlooking important aspects. Since a complete eDSR paper would exceed the scope of this paper, we focus on the routing algorithm and its outputs. Additional echelons derived from the requirements analysis, e.g. UI design (E2.1 in Fig. 1), are not covered in the design process here, and will be reported separately.

To strengthen our problem understanding (E1) and to derive initial requirements (E2), we first conducted a literature review on the topic of elderly mobility. To design a routing algorithm that is considerate of senior citizens, we decided on a participatory approach within one mid-sized German town, and, in addition, conducted semi-structured exploratory interviews with eleven older adults (8 female, 3 male; average age 68.3; U-01–11; 44 min. average) (E2). The semi-structured interviews followed a standardized guide derived from prior research on age-friendly mobility and navigation. Topics covered everyday activities and mobility contexts, technology use, perceived environmental and accessibility barriers, as well as route-level considerations such as effort-distance trade-offs, safety, comfort, and environmental quality. We analyzed the transcribed interviews using thematic analysis [5] to produce an initial set of DRs and DPs.

Based on the interview-derived design knowledge, we consolidated available data as well as data sources that enabled us to operationalize requirements. This scoping step resulted in a set of algorithm-centered design knowledge which were able to be instantiated within the route algorithm through specific route related data (E3). The algorithmic artefact constitutes a first functional implementation of senior-friendly routing, operationalizing senior-relevant constraints and supportive factors (E4). In the initial evaluation stage, the route algorithm was subjected to rigorous scrutiny to ascertain the validity of route selection and the properties of the routes. The development team undertook several routes on foot and checked each characteristic (E5).

In order to facilitate the evaluation of the route with the target group of older adults, a minimalistic mock-up user interface was created so that the one could use the algorithm in a feasible mobile application (E6). In order to facilitate an efficient evaluation process, a preliminary test (E7) was conducted. This test was designed to ascertain the basic walkability of the algorithm routes and the interaction with the mock-up interface. The preliminary test comprised three seniors (average age of 75.3 with two male and one female participant), who traversed the route initially designated as senior-friendly (SF) and subsequently returned via a commonly utilized Google Maps (GM) route, different from the SF route, to the initial starting point. Afterwards, the participants were interviewed for 15-20 minutes in semi-structured interviews, focused on the participants' perceptions of the routes and their preferences; the SF route was compared with the typical GM route. As the participants' feedback primarily concerned the presentation of the route on the UI, it was decided to modify the evaluation setup for the final evaluation cycle (E8) to refocus the participants on the route and its characteristics.

For our final evaluation cycle we conducted a field-based walking study that involved 14 older adults (9 female, 5 male; average age = 80.1 years, one missing age report; W-01–14; distinct from U-01–11). Walks were conducted in three groups of 2-7 participants, with each group having different starting and ending points and therefore following a different route pair. The walks were followed by interviews in groups of 1-4, with each session lasting about an hour. The standardized procedure comprised: a short introduction covering briefing, and basic demographic and mobility-related information; a guided walk (20 min in total; 800±100m per route) in which participants experienced two route alternatives, first the SF route and then the GM route, with no meaningful elevation differences between them; a brief demonstration of available route characteristics and supportive information (benches, toilets, shelters); and a semi-structured post-walk interview (17 min on average) eliciting perceptions of safety, comfort, physical effort, suitability for walking alone, and requirements for SF routing, like preferences for pre- versus in-walk information and route communication.

Additionally, participants provided brief quantitative assessments after experiencing both routes, rating safety, comfort/pleasantness, physical effort, and suitability for walking alone on 7-point Likert scales and indicating a comparative route preference. Quantitative data were analyzed descriptively to triangulate the qualitative findings.

To integrate evaluation results into the evolving design knowledge base, we initiated an additional design cycle that synthesizes findings from E8 with the design expertise developed in E2. This cycle served to confirm, refine, or extend existing DRs and DPs and, where needed, derive new ones. The resulting refined DRs and DPs subsequently informed the updated DPs in E10 and the corresponding prescriptive design knowledge.

Overall, the eDSR process concludes with a final synthesis that consolidates evaluation insights into actionable guidance for further artefact iterations and echelons.

# 4 The Development of Prescriptive Design Knowledge

## 4.1 Cycle 1: Initial Design Requirements, Instantiation, and Evaluation

Our literature synthesis (Section 2) informed the interview guide and sensitized us to feasibility constraints, safety thresholds, and amenity-driven comfort. The DRs and DPs reported here stem primarily from our empirical interviews and are linked to prior evidence where relevant. Underpinning participant quotes were translated by us.

Participants described heterogeneous mobility constraints and feasibility thresholds in their own lives or among peers that shape everyday walking decisions [cf. 1, 42]. Importantly, participants emphasized that mobility limitations do not imply inactivity, and framed going out as meaningful for participation and well-being [cf. 43]: "*you've got to speak to someone again*" (U-01); "*we are social beings*" (U-02).

Participants repeatedly described infrastructure barriers as route-defining constraints. Steep terrain and stairs were experienced as limiting or unacceptable (*"Up the mountain is difficult"* (U-02); *"where's an alternative? There are stairs"* (U-08)). This supports barrier-aware routing that treats obstacles and demanding segments as exclusions or penalties, aligning with barrier and fall-risk evidence in the literature [e.g., 19] and accessibility-aware routing approaches [28], motivating DR#01 (Table 2). Supportive amenities were framed as feasibility-enabling, not optional. Participants emphasized the need for usable benches (*"Every now and then a bench"* (U-02)) and described rest opportunities as necessary when routes become demanding (*"[Climbing* *stairs] there needs to be somewhere to sit"* (U-01)). Toilets emerged as critical and often urgent (*"Toilets are [...] fundamentally important"* (U-01); *"I really need them"* (U-10)). Participants also favored shelters along the route when it rains (*"So rain and a walker are ... two things that don't go together."* (U-09)). These findings align with evidence that amenities such as benches and toilets extend manageable walking range and reduce uncertainty [e.g., 18], especially valued under health constraints [17], motivating amenity-aware routing (DR#02). As participants simultaneously required barrier avoidance and enabling infrastructure, routing must implement dual logic: treat barriers as hard constraints or weighted penalties while optimizing for supportive amenities that keep routes feasible. This operationalizes older adults' route choice as sequential constraint satisfaction where accessibility and safety set thresholds, and amenities support manageability once thresholds are met [1, 12], leading to DP#01.

Since routes computed with this dual logic involve trade-offs between barrier avoidance and amenity optimization, we reasoned that users would benefit from understanding these characteristics before committing to a route [cf. 8, 35]. Participants asked for information that supports realistic planning beyond distance and time. One participant explicitly proposed difficulty grading to enable choice (*"Simply divide into difficulty levels"* (U-01)). Drawing on evidence that older adults particularly value transparency about algorithmic decision-making processes for building trust in AI systems [8], and that complex multi-criteria routing benefits from explainability [37], we designed DR#03 to provide simple summaries of execution-relevant characteristics: barrier presence, amenity availability, and terrain type. The system outputs interpretable route information (DP#02) that communicates criteria-based rationales, supporting realistic planning as multi-criteria routing introduces trade-offs.

**Table 2.** The initial design requirements and design principles.

<table>
<tr><td>DR#01 – Barrier-free routing for mobility limitations: If users have physical impairments or mobility restrictions, the system should be able to find barrier-free routes and suggest them as potential routes.</td><td rowspan="2">DP#01 – Balance exclusion constraints with enabling support in route computation: Design the routing algorithm to implement dual logic that treats barriers as hard or weighted exclusions while simultaneously optimizing for the presence and distribution of supportive amenities, ensuring routes are both physically feasible and psychologically manageable for older adults with varying capabilities and preferences.</td></tr>
<tr><td>DR#02 – Amenity-aware routing with distribution preferences: The routing algorithm should incorporate supportive amenities as route optimization factors, favoring routes with evenly distributed support opportunities over routes with clustered or absent amenities, and providing graceful degradation when constraints cannot be fully met.</td></tr>
<tr><td>DR#03 – Structured route characteristic outputs with execution-relevant information: When the routing algorithm proposes routes, it should provide structured outputs that summarize execution-relevant route characteristics derived from computed route properties (e.g., distance, steep segments, barrier presence, and amenity availability), including warnings when supportive infrastructure is sparse or when constraints cannot be fully met.</td><td>DP#02 – Make route recommendations interpretable: Design the routing system to output standardized "route card" summaries that translate computed route characteristics into interpretable information for users, supporting realistic planning by showing key route demands and relevant support opportunities, rather than presenting only geometry, distance, and time.</td></tr>
</table>

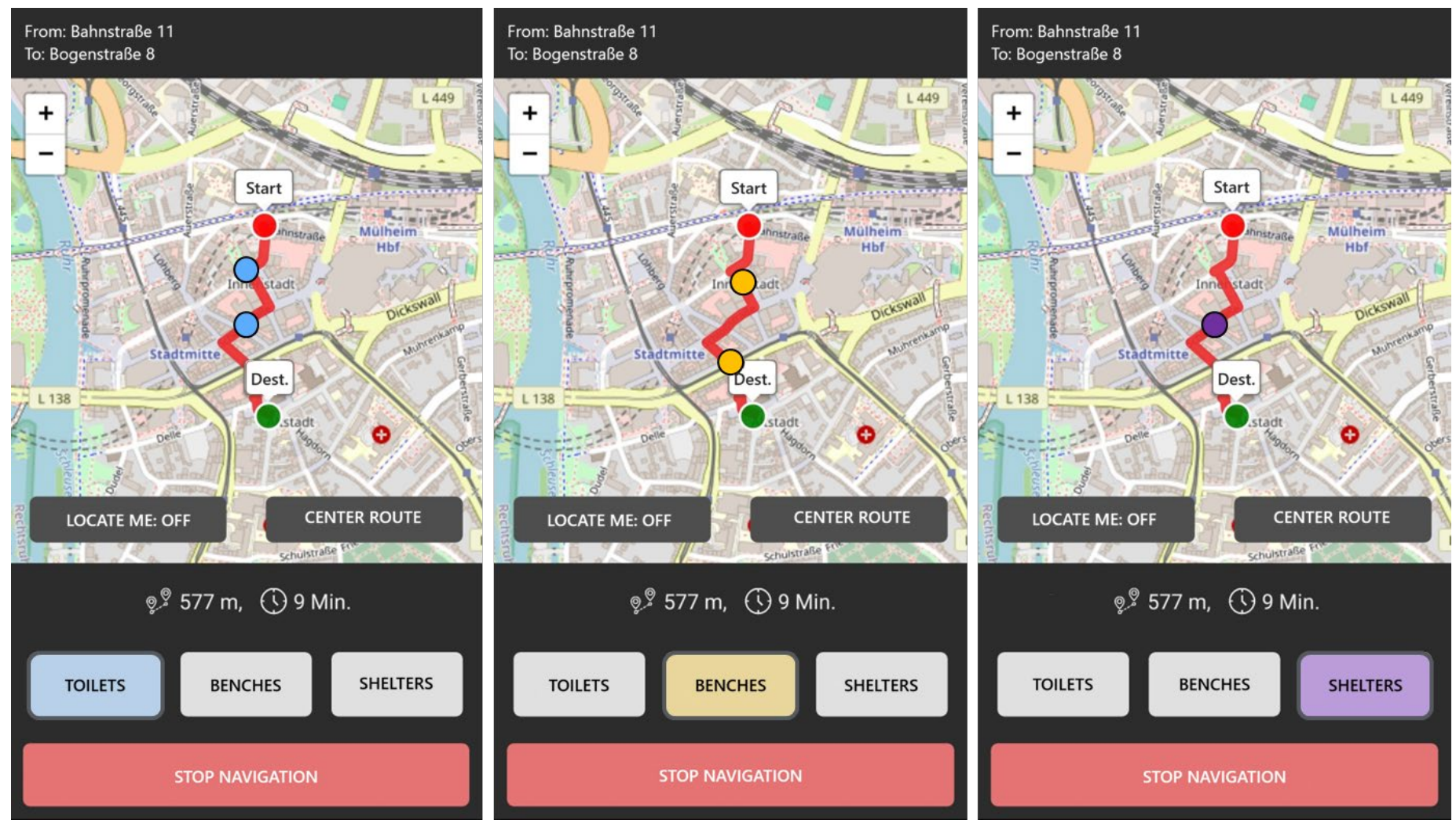

**Fig. 2.** Minimal demonstrator interface showing route and amenities along it.

**Artefact Instantiation:** To operationalize these DPs, we instantiated a routing artefact for one mid-sized German city, Mülheim an der Ruhr, comprising three integrated components: (1) an enriched pedestrian network graph, (2) a custom routing algorithm, and (3) structured interpretation payloads. The artefact's preprocessing, routing, and app code are available in a GitHub repository (www.github.com/SSKI-RUB/KISS).

We compiled the pedestrian network from OpenStreetMap (OSM), retaining senior-relevant attributes (surface type, stairs). We enriched each edge with elevation data from GEOportal.NRW to compute slope gradients and mapped addresses from Open-Addresses to intersections. Critically, we modeled amenities (benches, toilets, shelters) from OSM as amenities / points of interest (POIs) and pre-processed an accessibility graph connecting each POI to nearby network intersections, storing actual walk distances to enable efficient amenity-aware routing without runtime spatial computations.

The routing core operationalizes DP#01's dual logic through integrated barrier avoidance and amenity optimization. The algorithm implements A*-based path search with a configurable multi-criteria cost function (all weights, thresholds, and exclusions adaptable to individual constraints) incorporating: slope-category-based weighting with separate uphill/downhill penalties across multiple categories; surface-type weighting penalizing unpaved or gravel surfaces; hard exclusion rules for barriers; and amenity-aware optimization through dynamic POI bonus calculations. During graph traversal, the algorithm evaluates accessible amenities at each intersection, calculates bonuses based on POI type importance and distance decay, and adjusts route costs to favor paths passing by required amenities while preventing excessive detours via bounded bonuses.

The algorithm implements three-tier graceful degradation when constraints conflict, balancing feasibility with support (DP#01): routes with barrier-free paths and evenly distributed amenities (tier 1), routes with sufficient amenity counts but suboptimal spacing (tier 2), or barrier-optimized routes with incomplete amenity coverage (tier 3). Each tier generates warnings identifying unmet requirements, enhancing route interpretability (DP#02). Further operationalizing this DP, the artefact generates structured outputs translating computed characteristics into interpretable information: total and weighted distance, slope distribution, identified barriers, POI inventory with walk distances, and route quality classification with specific unmet requirements; extending beyond distance, and time to execution-relevant planning information. A minimal smartphone app displays computed routes and amenities (Fig. 2). While fully functional, it served solely to demonstrate algorithm outputs during evaluations.

**Pre-Evaluations: Data Validity and Pilot Feasibility.** Cycle 1 included two pre-evaluation steps that served as readiness gates for the field study. Evaluation 1 (E5) assessed whether the OSM-derived network and POI layers were sufficiently accurate to support a field-based route comparison. We conducted targeted ground-truth checks along the study area and candidate routes, verifying amenity existence/accessibility, network connectivity at decision points, and completeness/plausibility of key attributes used by the routing function (e.g., slope/surface labels, exclusions such as stairs). Minor discrepancies primarily affected POI completeness (e.g., unclear public-toilet labeling), but did not materially undermine the planned evaluation procedure. We therefore proceeded without changing routing logic but restricted the evaluation area and accounted for data uncertainty when interpreting results. Evaluation 2 (E7) checked feasibility and comprehension. A pilot session with three older adults indicated minor adjustments regarding route length, wording of questions, and demonstration procedure. Participants found short routes difficult to differentiate and articulated trade-offs more clearly when route exposure was longer. Thus, we doubled the route length in the final evaluation.

**Final Evaluation: Field-Based Route Comparison with Older Adults.** The final evaluation (E8) involved 14 older adults (9 female, 5 male; average age = 80.1 years, one missing age report). Participants experienced two routes (SF vs. GM) during a guided walk (20 min total; 800±100m per route), followed by a brief demonstration of route characteristics and supportive information, post-walk ratings, and a semi-structured post-walk interview. Participants rated *perceived safety*, *comfort/pleasantness*, *physical effort*, and *suitability for walking alone* on a 7-point Likert-scale (n=14; reported descriptively to triangulate qualitative insights). Across all dimensions, the SF route received higher mean ratings than the baseline (Table 3). In a comparative preference question, 9 participants (64%) preferred the SF route (6 *"clearly SF"*, 3 *"rather SF"*), 1 preferred the GM route (*"rather GM"*), and 4 (29%) reported no preference.

Post-walk interviews revealed that participants evaluated routes through infrastructure-maintenance interactions rather than just distance, aligning with evidence that micro-scale surface conditions disproportionately affect older walkers' fall risk and route acceptance [6, 33]. The SF route was valued for openness and calm (*"very pleasant because it was wider and open,"* (W-04)), yet seasonal concerns emerged as safety-defining: *"the leaves, with crutches you slip"* (W-02), *"in winter, the [GM] route is safer, maintained"* (W-04), validating that perceived hazards shift dynamically. Traffic exposure produced comfort-safety trade-offs. While the GM route was *"better swept,* *better looked after"* (W-03), traffic volume deterred participants: *"This traffic, I would rather walk the [SF] route"* (W-02). Others found pedestrianized segments *"more relaxed, no cars"* (W-10), yet balance concerns intensified where traffic and narrow footways converged: *"pay attention that you stay on your sidewalk"* (W-10). This reflects evidence that high-traffic streets are tolerated when necessary but avoided when alternatives exist [4, 33], connecting to sensory overload [23]. Narrow segments with obstructions reduced maneuverability (*"so tight, cars are parked there,"* (W-06)), and social-context sensitivity emerged: fear of walking alone in unpopulated areas (*"no people around,"* (W-05); *"I'm afraid, if I fall, there's no one around. People rarely pass by there"* (W-12)) underscores how visibility and surveillance shape perceived safety independent of infrastructure quality [33]. These patterns confirm that route acceptance reflects context-dependent constraint satisfaction, not fixed preferences.

The interviews also revealed recurring novel themes on how older adults perceive route differences and what they expect from routing support, which Cycle 2 details.

**Table 3.** Quantitative results of the route comparison: SF route / GM route (Δ).

| **Perceived safety** | **Comfort / pleasantness** | **Physical load** | **Suitability for walking alone** |
|---|---|---|---|
| 6.57 / 5.36 (17.35%) | 5.79 / 4.93 (12.24%) | 6.36 / 5.64 (10.20%) | 5.93 / 5.43 (7.14%) |

### 4.2 Cycle 2: Evaluation-Driven Refinement of Prescriptive Design Knowledge

While Cycle 1 established initial design knowledge (DR#01–03, DP#01–02) and instantiated the routing artefact, Cycle 2 refines and extends this foundation through field-based evaluation insights. The walking study (E8) first validated and refined the relevance of Cycle 1 requirements before revealing additional design considerations.

The field evaluation confirmed that barrier avoidance (DR#01) and amenity support (DR#02) are central to older adults' route acceptance. Participants consistently emphasized hazard triggers that align with Cycle 1's barrier focus: *"Fewer crossings would also be good. At least fewer traffic lights. Maybe more crosswalks"* (W-08); *"And there's often stuff lying around, because as a visually impaired person, when there's a scooter or something like that lying around, it's sometimes a problem"* (W-03). Regarding amenities, participants framed supportive infrastructure as enabling rather than optional: *"Benches. And a hint where the toilet is"* (W-04); *"Shelters wouldn't be bad either if the weather turns worse"* (W-08); *"Lighting is important"* (W-02). These findings validate that the dual routing logic operationalized in Cycle 1 (DP#01) addresses salient route-choice factors. Most participants welcomed explanations of route recommendations, particularly when grounded in senior-relevant criteria. For example: *"It would be important to communicate why"* (W-04), validating DR#03's focus on providing interpretable route information with criteria-based rationales (DP#02).

While Cycle 1 identified static barriers to avoid (DR#01), the evaluation revealed that perceived risk is situational. The same route segment is evaluated differently depending on weather conditions, lighting, and time of day. Participants noted that leaves create slip risk in autumn, that cleared main roads feel safer in winter, and that darkness transforms acceptable paths into no-go zones: *"I wouldn't go there in the dark."*, *"No* *way!"* (W-06 & W-07). This context-sensitivity is consistent with evidence that environmental perceptions interact with personal vulnerability and situational factors [19, 33]. We therefore extend barrier awareness toward context-sensitive hazard modeling (DR#04, Table 4) and consequently refine the barrier avoidance component of DP#01 (Cycle 1) by introducing situational modifiers, leading to the refinement of DP#01.

Participants did not seek a single "optimal" route but reasoned through context-dependent trade-offs: *"Or shelters. If I look out the window and say, oh, I can still run that distance before it starts raining."* (W-04). This finding aligns with evidence that older adults balance multiple criteria depending on in-situ conditions [12, 17]. This extends Cycle 1's single-route explanation payload (DR#03, DP#02) toward multi-route recommendations (DR#05). DR#05 refines DP#02 by extending the explanation payload from single-route characteristics to comparative multi-route information.

Several participants suggested landmark-based phrasing (*"Use landmarks like 'until the church and then right', that helps"* (W-03)), indicating preference for explanations anchored in recognizable environmental cues [cf. 10, 30], which also foster incidental spatial knowledge acquisition during navigation [45], especially relevant in unfamiliar spaces with confusing layouts and unclear signage [23]. This requires the algorithm to be aware of landmarks (DR#06). It addresses how route information is communicated by anchoring abstract algorithmic decisions in users' lived environment (DP#03).

Walking alone versus accompanied affected perceived tolerance for scenic but isolated segments: *"Here I feel the loneliness ... no people"* (W-07). This motivates social-context awareness (DR#07) and extends the routing algorithm's input parameters to include contextual factors beyond individual mobility constraints, enabling personalized recommendations that adapt to walking situations (DP#04).

Participants distinguished planning-time needs (overview of length, difficulty, amenity availability: *"How long is the route, and where is the next place I can buy water?"* (W-02)) from execution-time needs (timely warnings and nearby support). As routing becomes more safety-relevant, explanation content must support realistic execution [37]. This motivates stage-structured algorithm outputs (DR#08 & DR#09) and specifies the information architecture: routing outputs must be modular and stage-tagged to support different consumption patterns throughout the journey (DP#05).

Together, these refined DRs and DPs provide prescriptive guidance for advancing senior-friendly pedestrian navigation beyond distance or time optimization.

**Table 4.** The expanded and refined design requirements and design principles.

<table>
<tr><td>DR#01 - Barrier-free routing for mobility limitations (cf. Table 2).</td><td rowspan="3">DP#01 – Context-aware feasibility balancing: Design the routing algorithm to implement dual logic that treats barriers as context-dependent exclusions (adjusted by weather, lighting, time of day) while optimizing for amenity distribution, recognizing that route feasibility is situational rather than static.</td></tr>
<tr><td>DR#02 – Amenity-aware routing with distribution preferences (cf. Table 2).</td></tr>
<tr><td>DR#04 – Context-aware hazard sensitivity: The routing algorithm should in-corporate situational modifiers (weather conditions, lighting, time of day) that adjust barrier penalties and safety thresholds dynamically, recognizing that hazard salience varies with environmental context.</td></tr>
</table>

<table>
<tr><td>DR#03 – Structured route characteristic outputs with execution-relevant information (cf. Table 2).</td><td rowspan="2">DP#02 – Provide transparency through multi-route explanation payloads: Design the routing algorithm to output interpretable information for multiple route alternatives, communicating not just individual route characteristics but also comparative trade-offs, enabling informed situational choice rather than single-solution recommendations.</td></tr>
<tr><td>DR#05 – Multi-route recommendation with explicit trade-offs: The routing algorithm should generate multiple candidate routes representing different optimization positions (safest vs. shortest vs. most pleasant), with structured outputs that expose the relative strengths and weaknesses of each alternative.</td></tr>
<tr><td>DR#06 – Landmark-grounded route explanations: The routing algorithm should identify and incorporate recognizable landmarks along route segments to support explanation outputs that ground abstract attributes (distance, direction, surface conditions) in concrete environmental features, enabling spatial orientation.</td><td>DP#03 – Ground algorithmic outputs in environmental landmarks: Design the routing algorithm to identify and tag recognizable landmarks along computed routes, enabling explanation outputs that reference familiar environmental features rather than only abstract directions, supporting both orientation and comprehension.</td></tr>
<tr><td>DR#07 – Social-context sensitivity in route recommendations: The routing algorithm should accept social context parameters (walking alone vs. accompanied) as input and adjust route recommendations accordingly, recognizing that companionship affects perceived safety thresholds and preference for predictable versus exploratory routes.</td><td>DP#04 – Incorporate social context in route personalization: Design the routing algorithm to accept social context parameters and adjust route selection accordingly, recognizing that companionship modulates both perceived safety thresholds and preference for predictable versus scenic routes.</td></tr>
<tr><td>DR#08 – Planning-time route overview with execution-relevant detail: The routing algorithm should generate planning-time outputs that summarize route length, expected difficulty/effort, and the availability and distribution of critical amenities (benches, toilets, water/food access), enabling feasibility assessment.</td><td rowspan="2">DP#05 – Structure route outputs as stage-appropriate information payloads: Design the routing algorithm to generate structured outputs that distinguish between planning-time information (comprehensive route characteristics, amenity distribution, difficulty assessment) and execution-time information (event-triggered cues about upcoming amenities, hazards, decision points), enabling adaptive information consumption throughout the journey.</td></tr>
<tr><td>DR#09 – Execution-stage event-triggered information: The routing algorithm should identify and tag execution-relevant events (upcoming amenities, hazard warnings, decision points) so that appropriate cues can be delivered at the right time during route execution.</td></tr>
</table>

# 5 Discussion

Our primary contribution is the systematic derivation and empirical refinement of design knowledge that extends beyond static accessibility modeling, and answers our research question. The refined DP#01 advances prior work by recognizing that barrier salience is situational rather than fixed; what constitutes an acceptable route shifts with weather, lighting, and time of day. This finding challenges the common assumption in accessibility routing that barriers can be modeled as static attributes and suggests that effective senior-friendly systems require dynamic context inputs. The refined DP#02 addresses a gap in current practice where pedestrian routing systems typically present single solutions. Our evaluation showed that older adults engage in context-dependent trade-off reasoning rather than seeking universal solutions, suggesting that transparency mechanisms should enable genuine decision-making rather than merely building trust in algorithmic recommendations. DP#03 operationalizes a preference repeatedly expressed by participants: algorithmic decisions should be anchored in recognizable environmental features rather than abstract directions or coordinates. This extends from purely algorithmic transparency toward spatial-cognitive grounding. DP#04 and recognizes that companionship modulates safety thresholds. DP#05 shows that information needs differ fundamentally between planning and execution stages. The DPs are formulated at a level of abstraction that should transfer across contexts, though specific instantiation parameters require local adaptation. The logic, context-aware barrier avoidance, amenity distribution optimization, multi-route transparency, landmark-based communication, social-context sensitivity, and stage-structured information, represents guidance for developers of senior-oriented mobility support systems.

Our evaluation was conducted in one mid-sized German city, limiting empirical validation in different urban forms, climates, and cultural contexts. The design knowledge remains valid as prescriptive guidance, but instantiation parameters and their effectiveness need context-specific validation. Group-based walking may have influenced participants' perceptions compared to individual use. Although participants did not know the interview questions in advance, interaction during the walk may still have influenced how route impressions were articulated in the interview; at the same time, such exchange also helped participants reflect on and verbalize route characteristics. Some participants may have been generally familiar with the city, although not necessarily with the exact study routes. Some DRs remain specifications rather than implemented features. These limitations define boundary conditions for knowledge claims rather than undermining the prescriptive value of the DRs and DPs.

Future work should evaluate the full set of refined DRs: context-aware hazard modeling, multi-route generation, landmark identification, social-context parameters, and stage-appropriate information architecture. Incorporating additional network parameters/amenities would also be valueable. Our work also speaks to the paradox of prevention: systems aimed at avoiding adverse outcomes face the challenge that users may not perceive their value until harm has already occurred [40]. Senior-friendly routing is no exception: its benefits for fall avoidance and sustained participation may only become apparent retrospectively, so future iterations should attend explicitly to communicating preventive value to users. A further promising direction is feedback-based personalization. The high configurability of our routing algorithm provides a basis for incorporating repeated user feedback to iteratively adapt route weights and constraints, enabling more fine-grained personalization. Long-time field studies would validate whether the design knowledge translates into sustained mobility benefits. Future research should also examine how cultural context and local living conditions shape older adults' walking practices, safety perceptions, and the acceptance of senior-friendly routing recommendations. Factors like neighborhood structure, socioeconomic conditions, and cultural norms may shift both relevant barriers and the relative importance of route features [12]. Future echelons require parallel development and integration into overall systems.

## 6 Conclusion

Older adults' independent mobility and the social participation it enables protect physical, cognitive, and mental health in later life. When mobility declines, older adults progressively withdraw from outdoor activity, risking a self-reinforcing cycle of social isolation, deconditioning, and diminished well-being. Technologies that reduce barriers to going out can help interrupt this cycle. Yet mainstream pedestrian navigation systems optimize for distance or time while overlooking the environmental barriers, safety thresholds, and supportive infrastructure that shape late-life walking decisions.

We address this gap through an eDSR approach, deriving and refining prescriptive design knowledge grounded in older adults' lived mobility constraints. Through eleven semi-structured interviews, we derived initial DRs and DPs, which we instantiate in a routing artefact that generates personalized, barrier-aware, amenity-sensitive routes with structured explanation payloads. Field evaluation with 14 seniors validated that artefact-generated routes were more senior-friendly than the GM baseline. Finally, we contribute nine DRs and five refined DPs emphasizing context-aware barrier avoidance with amenity optimization (DP1), multi-route transparency with trade-off explanations (DP2), landmark-grounded communication (DP3), social-context sensitivity (DP4), and stage-appropriate information architecture (DP5).

These contributions provide actionable guidance for practitioners developing navigation systems that support safe, confident, and autonomy-preserving mobility in later life, helping older adults sustain the social participation that protects well-being.

**Acknowledgments.** This study was funded by the German Federal Ministry for Education, Family Affairs, Senior Citizens, Women and Youth (grant number 3923406K07).

**Disclosure of Interests.** The authors have no competing interests to declare that are relevant to the content of this article.

*Declaration of AI: During the preparation of this work, the authors used Claude 4.5 Opus to proofread sections of the manuscript and assist with rephrasing selected sentences.*